\documentclass[10pt, a4paper]{article}
\usepackage{diagbox}
\usepackage[final]{lrec2026} 
\usepackage{tikz}
\usetikzlibrary{arrows.meta, positioning}
\usetikzlibrary{calc} 
\usepackage{booktabs}
\usepackage{graphicx}
\usepackage{diagbox}
\usepackage[most]{tcolorbox}
\usepackage{xcolor}
\usepackage{multirow}
\definecolor{myred}{RGB}{178,34,34}
\definecolor{vermeerblue}{RGB}{83,103,162}   
\definecolor{vermeeryellow}{HTML}{F1C27D} 

\newcommand{\nk}[1]{\textcolor{teal}{#1}}
\newcommand{\jp}[1]{\textcolor{black}{#1}}

\tcbset{
  prompt/.style={
    colback=vermeerblue!15,      
    colframe=vermeerblue,        
    colbacktitle=vermeerblue,    
    coltitle=white,              
    fonttitle=\bfseries,
    title=Prompt
  },
  response/.style={
    colback=vermeeryellow!20,    
    colframe=vermeeryellow,      
    colbacktitle=vermeeryellow,  
    coltitle=white,              
    fonttitle=\bfseries,
    title=Response
  }
}
\definecolor{arrowbrown}{HTML}{5F6946}
\usepackage{stfloats}

\title{Distributed Partial Information Puzzles: Examining Common Ground Construction Under Epistemic Asymmetry}

\name{
\begin{tabular}{c}
Yifan Zhu$^1$, Mariah Bradford$^2$, Kenneth Lai$^1$, Timothy Obiso$^1$ \\
Videep Venkatesha$^2$, James Pustejovsky$^1$, and Nikhil Krishnaswamy$^2$
\end{tabular}
}
\address{$^1$Brandeis University, $^2$Colorado State University \\
         415 South St, Waltham, MA 02453, Fort Collins, CO 80523 \\
         \{zhuyifan, jamesp\}@brandeis.edu,\\
         mbrad@rams.colostate.edu, nkrishna@colostate.edu}

\abstract{
Establishing {\it common ground}, a shared set of beliefs and mutually recognized facts, is fundamental to collaboration, yet remains a challenge for current AI systems, especially in multimodal, multiparty settings, where the collaborators bring different information to the table. We introduce the {\bf Distributed Partial Information Puzzle (DPIP)}, a collaborative construction task that elicits rich multimodal communication under epistemic asymmetry. 
We present a multimodal dataset of these interactions, annotated and temporally aligned across speech, gesture, and action modalities to support reasoning over propositional content and belief dynamics.
We then evaluate two paradigms for modeling common ground (CG): \textcolor{black}{(1) state-of-the-art large language models (LLMs), prompted to infer shared beliefs from multimodal updates, and (2) an axiomatic pipeline grounded in Dynamic Epistemic Logic (DEL) that incrementally performs the same task.} 
Results on the annotated DPIP data indicate that it poses a challenge to modern LLMs' abilities to track both task progression and belief state.
 \\ \newline  }

\begin{document}

\maketitleabstract

\section{Introduction}
\label{sec:intro}

Effective collaboration between humans depends on establishing {\it common ground}---a set of shared beliefs and agreed-upon facts in a task context. Doing so enables them to align perspectives, coordinate actions, and achieve joint goals \cite{clark_grounding_1991,traum1994computational,asher1998common,dillenbourg2006sharing}. However, humans rarely begin a task with the exact same background, perspective, or information, meaning that construction of common ground requires them to resolve {\it epistemic asymmetry}~\cite{zhou-etal-2024-real} and effectively communicate and make inferences about partial or privately-held information that their interlocutors may have. These complexities are amplified in {\it co-situated interaction}, which involves significant multimodal components, as humans communicate through both verbal cues (speech) and non-verbal cues such as gestures and actions in context~\cite{cassell2000embodied,wahlster2006dialogue,foster2007enhancing,kopp2010gesture,marshall2013theories,schaffer2019conversation}. The challenges are multiplied further when collaborations involve more than just two parties, as in this case each individual has to integrate and coordinate with multiple information sources. Common ground, dialogue state tracking, collaboration, theory of mind, and related issues have been the objects of intense study and interest within dialogue research, NLP, and AI more generally. However, data that enables the robust study of these topics in multiparty, co-situated, epistemic-asymmetric collaboration, and illuminates the challenges it poses to state-of-the-art (SOTA) AI systems remains sparse.

\begin{figure}
    \centering
    \includegraphics[width=\linewidth,trim={0px 100px 0px 100px},clip]{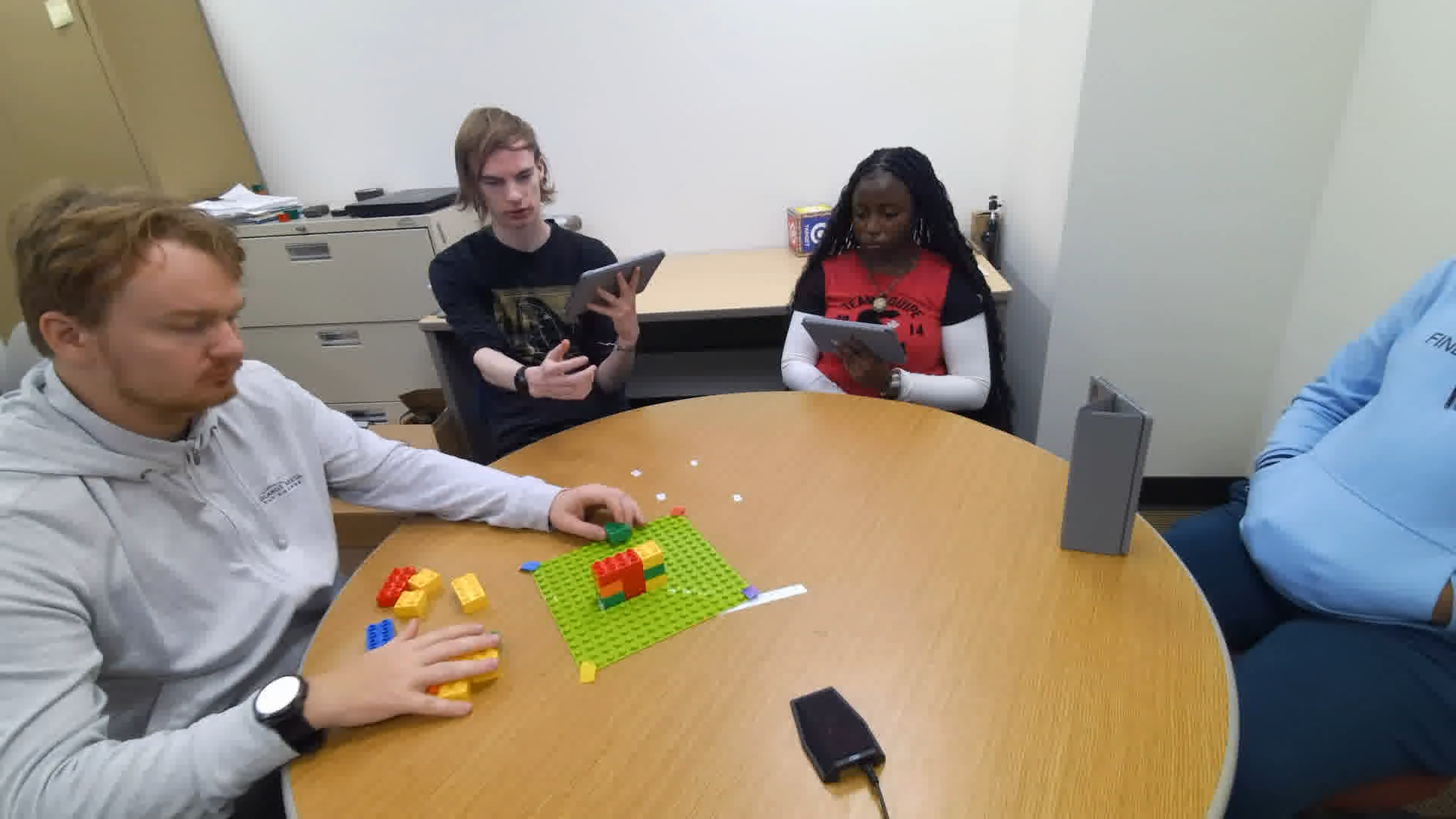}
    \vspace*{-2mm}
    \caption{A \textit{builder} and 3 \textit{directors} participating in the DPIP task with a partially-completed structure on the table. Director 1 (second from left) is indicating the position of a block using a gesture as well as a speech act.}
    \label{fig: group example}
    \vspace*{-4mm}
\end{figure}

In this paper, we introduce a task we call a ``Distributed Partial Information Puzzle'' (DPIP), which is realized as a type of collaborative construction task. In the task, three ``directors'' are each given individual partial information about a goal structure and must collaboratively instruct one ``builder'' to construct a single contiguous structure that is that consistent with all three pieces of individual information (Fig.~\ref{fig: group example}). We present a dataset of DPIP interactions annotated with speech transcriptions, gestures, and actions in context, as well as the propositional information communicated through each of these modalities. Further, we investigate the ability of SOTA LLMs to track and reason about information exchange in this context, to establish the challenges this type of task and domain poses to SOTA AI systems.  Our annotated data and evaluation code, including prompts, are available through \url{https://doi.org/10.5281/zenodo.18626419}, and will be a continually updated community resource.

\section{Related Work}
\label{sec:relatedwork}

\textit{Common ground} has long been studied in linguistics, philosophy, and computational dialogue research as the process by which interlocutors establish shared understanding through communicative acts \cite{clark_grounding_1991,traum1994computational}. Later work formalizes its role within discourse structure and collaborative learning \cite{asher1998common,dillenbourg2006sharing}, viewing common ground as an evolving set of shared beliefs that enable coordination and joint action.

Because human communication is inherently multimodal, grounding extends beyond speech. Prosody conveys intention beyond propositional content \cite{shriberg2004prosody,chiu2008flowing}, gaze signals attentional focus and shared reference \cite{fathi2012learning,huang2015using}, and gesture complements or substitutes for verbal acts to enrich meaning \cite{komorowska2017speech,brutti-etal-2022-abstract}. Together these modalities form the substrate through which common ground dynamically emerges.

Situated language understanding has been explored previously in physical and virtual environments \cite{zarriess2016pentoref,krishnaswamy2017communicating,pustejovsky2017creating,krishnaswamy2018evaluation,bisk2018learning,chen2019touchdown,suhr2019executing,krishnaswamy2020diana,krishnaswamy2022voxworld,thomason2020jointly}. Related work models how interlocutors infer intentions and beliefs \cite{chai2014collaborative,imai2025measuring}, track dialogue state \cite{williams2016dialog}, and display theory-of-mind–like reasoning \cite{sap2022neural,ullman2023large,nath-etal-2025-frictional,sicilia2025evaluating}. However, few datasets capture all three challenges---multiparty interaction, co-situatedness, and epistemic asymmetry---simultaneously.

Among existing corpora, the HCRC Map Task \cite{anderson1991hcrc} and EGGNOG \cite{wang2017eggnog} model two-party co-situated collaboration; \textit{MindCraft} \cite{bara-etal-2021-mindcraft} introduces partial observability in a virtual world; and \textit{DeliData} \cite{karadzhov2023delidata} provides multiparty deliberation but no embodied action. The Weights Task \cite{khebour2024text,khebour2024commongroundtrackingmultimodal,khebour2025feature,vanderhoeven2025trace} combines gesture and action in a co-situated setting, yet its full observability constrains the epistemic richness of belief negotiation. Our \textit{DPIP} task complements these efforts by uniting all three aspects—multimodality, multi-party interaction, and asymmetric knowledge—within a collaborative construction domain.

Work on competitive or deceptive multi-agent settings also examines reasoning under partial information. For example, \citet{zhang2025multimind} develop \textit{Werewolf} agents that infer hidden roles and intentions, extending theory-of-mind frameworks \cite{baron1997mindblindness,goldman2012theory,pustejovsky2021embodied}. In contrast, the DPIP task is cooperative: participants integrate truthful but partial evidence to build a shared representation of a physical goal. While \citet{zhang2025multimind} emphasize emotional and deceptive cues, our focus is on \textit{task-oriented multimodal grounding}, where gesture and action encode propositional information central to problem-solving.

Complementary work by \citet{sileo2023mindgames} employs Dynamic Epistemic Logic (DEL) to evaluate theory-of-mind reasoning in text-based scenarios. Our approach extends this logic-driven perspective to a co-situated multimodal setting, where updates to belief states must be inferred jointly from speech, gesture, and action. Large language models (LLMs) show strong results in discourse reasoning and implicit inference \cite{fei2024multimodal,niu2024enhancing,zhang2025scenellm}, yet to our knowledge, no prior study has systematically evaluated whether LLMs can infer and track \textit{common ground} in collaborative, multimodal, and partially-observable environments such as DPIP.

\section{Task Description and Dataset}
\label{sec:data}

\subsection{Distributed Partial Information Puzzle}
\label{ssec:dpip}

The {\it Distributed Partial Information Puzzle} (DPIP), is a collaborative problem-solving paradigm in which essential information is \textit{distributed} among participants, making communication a requirement for task success. Unlike traditional shared-information tasks, DPIPs deliberately prevent any single participant from solving the puzzle independently, thereby enforcing communication as the primary means of coordination.

Our DPIP task is a collaborative construction task, performed in groups of four, wherein each of three ``directors'' is given different partial information about a structure and have to collaboratively instruct one ``builder'' to build a single contiguous structure out of large Lego blocks, that is consistent with all pieces of information. See Fig.~\ref{fig: group example}. The task is complete when all three directors agree that the structure that the builder has built is consistent with all three of their views. The group is \textit{successful} if that structure actually matches the originally-generated side views.

\begin{figure*}
    \centering
    \includegraphics[width=.3\linewidth,trim={450px 230px 450px 140px},clip]{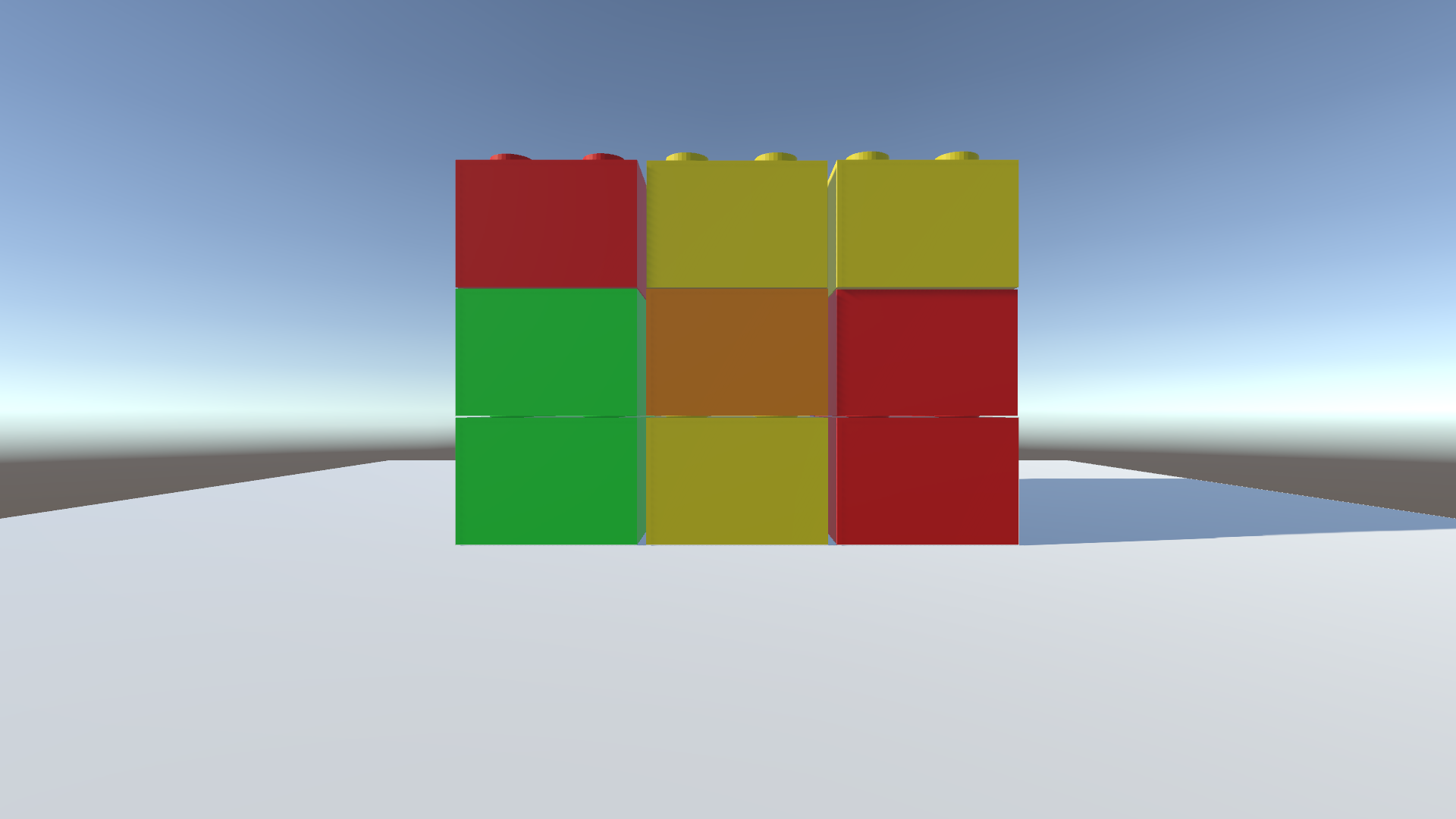}
    \includegraphics[width=.3\linewidth,trim={350px 200px 450px 100px},clip]{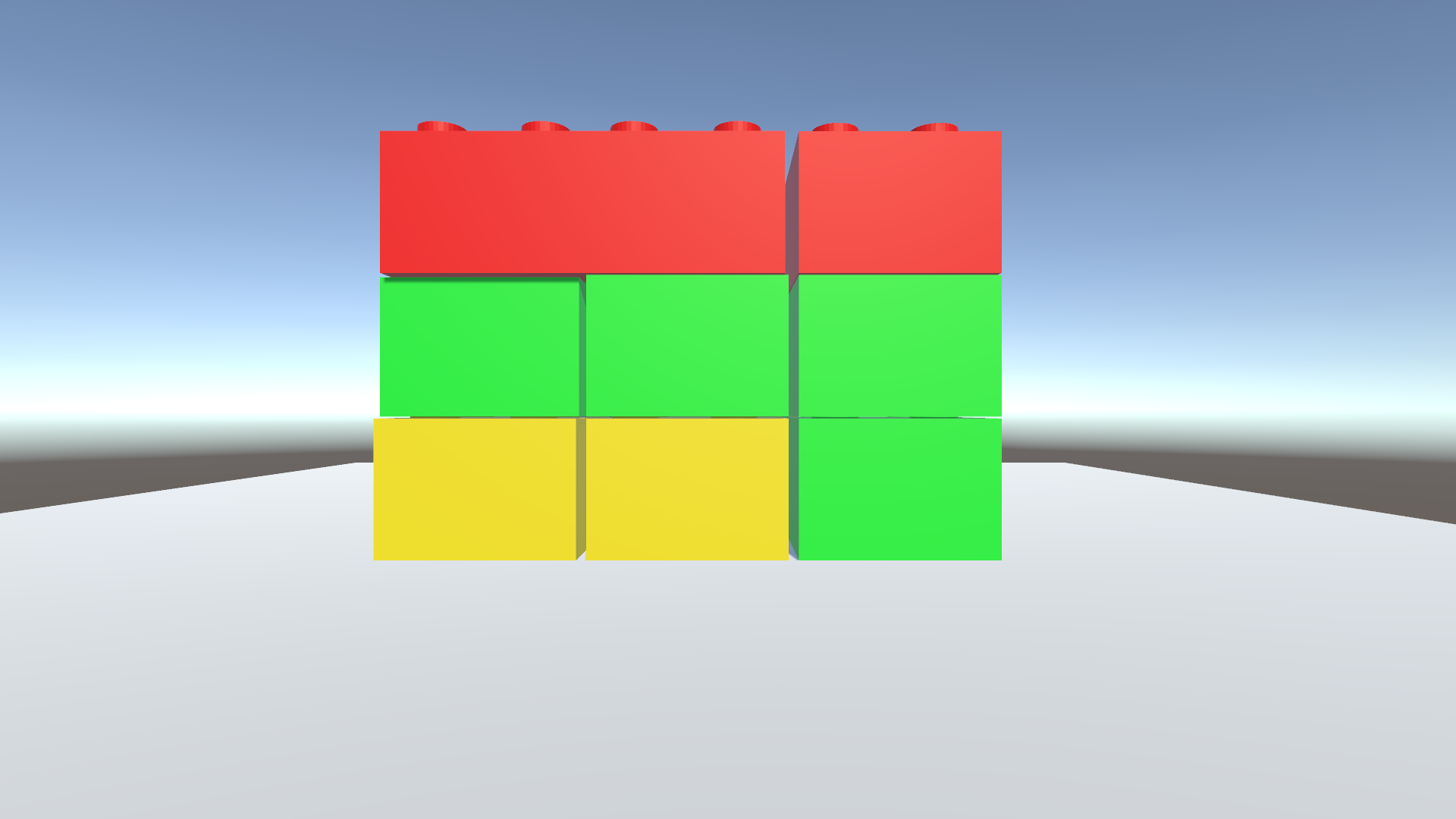}
    \includegraphics[width=.3\linewidth,trim={300px 200px 500px 100px},clip]{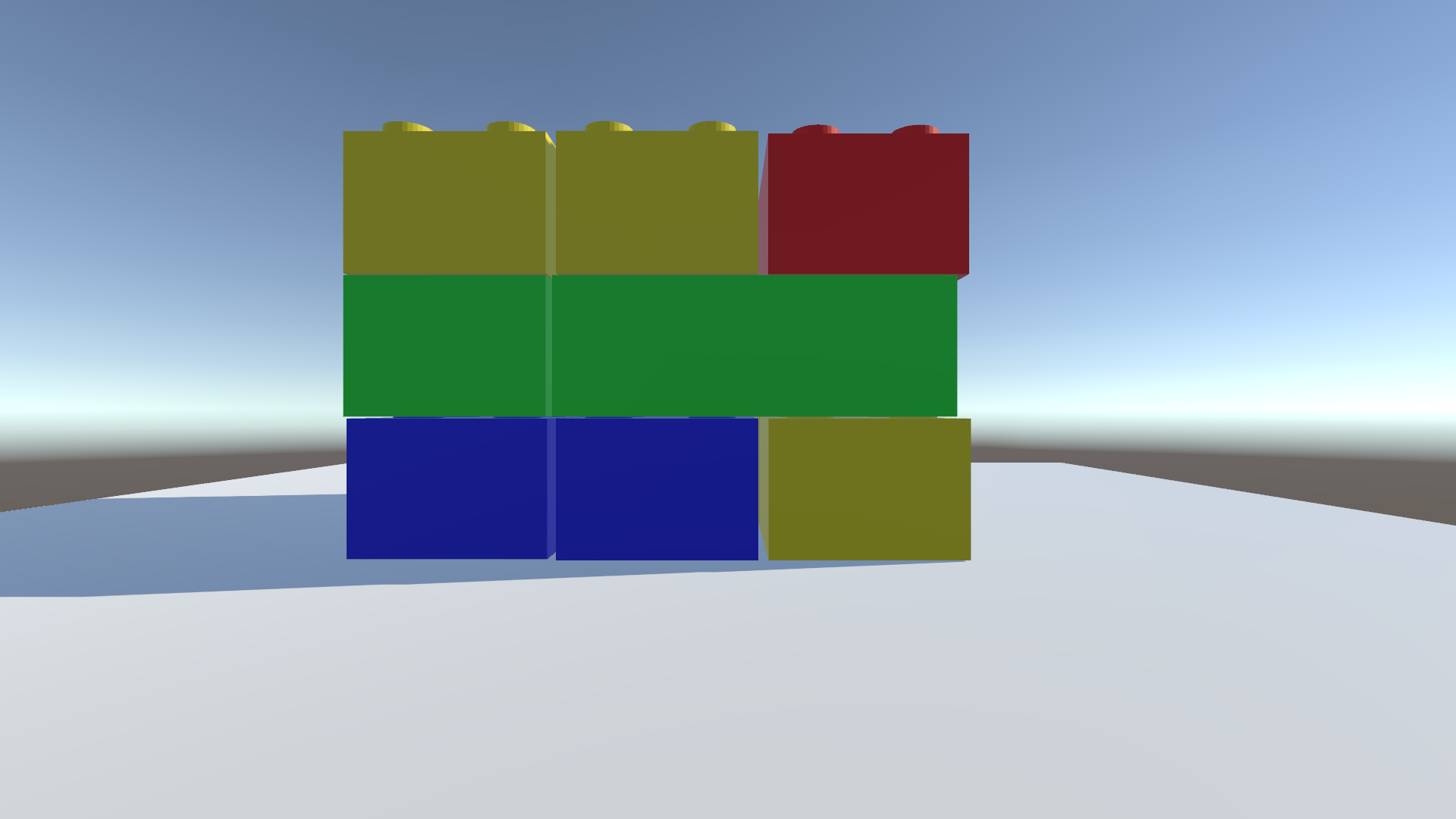}
    \vspace*{-2mm}
    \caption{3 individual side views of the same complete structure, each assigned to a director.}
    \vspace*{-4mm}
    \label{fig:sideviews}
\end{figure*}

In our data collection implementation, the goal structure has dimensions of $3~\text{(width)} \times3~\text{(depth)}\times3~\text{(height)}$, where the unit of the height dimension is a layer of blocks and the unit in the width and depth dimensions is {\it 2} of the pegs atop a Lego block (thus a single ``square'' Lego is considered to be $1\times1\times1$ --- see Fig.~\ref{fig:sideviews}). The structure is composed entirely of square or rectangular blocks, and there are no gaps in between blocks in any of the walls. In a variant of the task, the structure footprint is extended to $4\times4\times3$, and blocks with curved sides and gaps in the structure are allowed---however, our annotations and evaluation extend only to the first variant described above.

Epistemic asymmetry is established by distributing distinct 2D side views to the three directors, while only the builder is allowed to manipulate the blocks. The ground truth goal structure is procedurally computed and rendered in the Unity game engine, and screenshots of the walls (e.g., Fig.~\ref{fig:sideviews}) are distributed to the directors. The directors must collectively guide the builder to reconstruct the goal structure, and in the process must necessarily reconcile discrepancies across their partial perspectives, such as inferring when another participant is expressing or doing something that is consistent or inconsistent with their own private information. Each director holds unique spatial information, and as only the builder can touch the blocks, no group can complete the task without contributions from all members. This setup simulates teams with members with different backgrounds and expertise.

This setup enables the study of epistemic group states and common ground formation, as participants must externalize private visual knowledge through language and gestures (and in the case of the builder, actions). The task further demands precise spatial reasoning and description, and even mental simulation \cite{goldman2006simulating}, since the builder must form a mental model of the structure without visual reference.


\subsection{Data}
\label{ssec:data}

The annotated DPIP Lego dataset consists of 10 audiovisual recordings of groups of four performing the task described above (Fig.~\ref{fig: group example}). 
Each director was provided with a tablet showing a different side view of the target structure (e.g. front, left, or right side views---Fig.~\ref{fig:sideviews}) without knowing what side they are seeing. 
Participants were seated around a table and their interactions were captured using 3 Microsoft Kinect Azure cameras to log different angles of the task. Audio was recorded on a single conference-style tabletop microphone. The study was approved by university Institutional Review Boards (IRBs) and participants received USD 15.00 each.

Table~\ref{tab:descstats} provides descriptive statistics over the 10 groups. One of the groups, Group 7, was an extreme outlier in terms of task completion time ($Z$ score of 14.7), and did not successfully complete the task, and so we provide statistics with and without this group, which was held out from the main evaluation for a secondary experiment (Sec.~\ref{ssec:exp}). Of those subjects who completely filled out a demographic survey at study time, the average age was 23 years (SD = 3.58). Ten subjects were female and 21 were male. Eleven were Caucasian, 18 were Asian, and 2 reported other ethnicity. Fourteen were native English speakers, 11 were native Telugu speakers, 6 spoke other native languages.\footnote{Numbers are not reported for subjects who did not respond to questions on the demographic survey. Other native languages in the subject pool included Indonesian, Kannada, Malayalam, Spanish, Tamil, and Yoruba.} The task was conducted entirely in English.


\begin{table*}[h!]
    \centering
\begin{tabular}{lrrrrrrrr}
\toprule
 & Ave & SD & Ave* & SD* & Min & Max & Max* & Total \\
 \midrule
Task Time (s.) & 703.53 & 903.48 & 423.52 & 190.43 & 159.82 & 3223.60 & 800.40 & 7035.32 \\
\# Utterances & 311.20 & 397.90 & 190.33 & 117.34 & 38 & 1399 & 350 & 3112\\
\# Actions & 47.90 & 48.34 & 33.00 & 11.48 & 19 & 182 & 53 & 479 \\
\# Gestures & 27.80& 19.74 & 22.56 & 11.35 & 7 & 75 & 37 & 278\\
\# Props & 30.70& 15.54 & 26.44 & 8.23 & 18& 69 & 40 & 307 \\
\# Tokens & 1302.70& 1442.70 & 860.44 & 375.71 & 350 & 5283 & 1517 & 13027 \\
\bottomrule
\end{tabular}
    \caption{Summary statistics of 10 annotated DPIP Lego task groups. *Indicates statistics excluding outlier group (Sec.~\ref{ssec:data}).}
    \label{tab:descstats}
\end{table*}

\section{Annotation Pipeline}
\label{sec:anno}

As task-relevant propositional content, as well as epistemic positioning, may be communicated through any of three primary modalities---\textit{action}, \textit{speech}, and \textit{gesture}---these form the empirical basis for modeling multimodal, multiparty communication in the DPIP task. Following~\citet{vanderhoeven2025trace}, we treat gaze as an implicit perceptual cue, while gesture is considered a distinct communicative modality complementary to speech. Each constituent modality is dually-annotated by independent human annotators, and inter-annotator agreement (IAA) scores are computed to assess reliability. When performing the annotations, all human annotators received access to the video recordings of the task, specifically to the central angle of the 3 camera angles recorded. 
Our annotation extends the process initially described in \citet{zhu2025multimodal}.

\subsection{Speech Annotations}
\label{ssec:speech-anno}

For the speech modality, we employ the Whisper ASR model \cite{radford2023robust} for automatic transcription. The automatic transcriptions are verified and corrected by a human who reviewed the transcripts by watching the associated videos. Annotators then annotated the resulting post-corrected utterances with the {\it propositions} they express, in the form of relational information among blocks referenced in dialogue. 

Each resulting proposition encodes four elements: the \textit{timestamp}, \textit{speaker identity}, \textit{relative position} between two blocks, and the \textit{side} information corresponding to the director’s assigned perspective. 
For example, if Director 2 says, “a small blue block is next to a green block from my side,” the extracted proposition is represented as:
\texttt{(nextto(BlueShortBlock, GreenShortBlock),  D2’s side)}.

\subsection{Action Annotations}
\label{ssec:action-anno}

The action space of the DPIP Lego task is limited to block manipulations by the builder: \textit{put} (a block appears at a location on the board), \textit{remove} (a block disappears from a location), and \textit{move} (a sequential {\it remove}-{\it put} operation). Therefore to derive gold-standard action sequences, we first had annotators log the evolution of blocks on the board in a 3D Structure Annotation Tool (SAT). While watching the video, annotators reconstruct the evolving board state by placing color- and shape-specified blocks onto a grid. The SAT interface records these placements as JSON logs containing block identifiers, spatial coordinates, and timestamps, while a 3D visualization panel provides a real-time way to verify annotation against the video recording.


From these logs, we then deterministically extracted discrete \textit{put}/\textit{remove}/\textit{move} actions according to the above definitions. Locations are first encoded as absolute coordinates and subsequently transformed into task-relevant spatial relation predicates (e.g., $on$, $leftof$, etc.).

Each action annotation integrates four key elements: the \textit{timestamp}, \textit{action type}, a \textit{unique object identifier with its block attributes (color and shape)} automatically generated by the system, and \textit{side} information added by human annotators. For example, the annotation
$put(rs1,~on(base)),~D1’s~side$
indicates that a \texttt{red short block} with the identifier \texttt{1} (unique among blocks of the same properties) is introduced into the structure from \texttt{Director 1}’s perspective, placed on the base layer.

\subsection{Gesture Annotations}
\label{ssec:gesture-anno}

Participant gestures are annotated using Gesture Abstract Meaning Representation (GAMR;~\citet{brutti-etal-2022-abstract}), a framework designed to capture gesture semantics \textcolor{black}{in ELAN~\cite{wittenburg2006elan}}. Gestures may be deictic, iconic, or emblematic, indicating structural descriptions 
(e.g., \textit{side by side}), block attributes (e.g., \textit{square}, \textit{rectangular}, \textit{curved}), 
or actions (e.g., \textit{bring forward/backward}), showing epistemic position such as agreement (e.g., \textit{nodding}).

Annotators with previous expertise in GAMR watched the video while annotating, to ensure annotations were time-stamped for alignment with utterances and actions, and are produced with access to object IDs, allowing gestures that reference specific objects to be concretely recorded. Annotators recorded when they saw gestures occurring and recorded their meaning. For example, the following GAMR annotation:
\begin{verbatim}
(d / deixis-GA
    :ARG0 (d1 / director-1)
    :ARG1 (bs1 / blue-square-1)
    :ARG2 (g / group))
\end{verbatim}
indicates that Director 1 ({\tt ARG0}) is performing a pointing gesture toward a specific block, \texttt{blue-square-1} ({\tt ARG1}), with the intended referent of the gesture being the group ({\tt ARG2}). 


\subsection{Inter-Annotator Agreement}
\label{ssec:iaa}

\renewcommand{\arraystretch}{1.25}
\begin{table*}[t]
\centering
\resizebox{\textwidth}{!}{
\begin{tabular}{lccccccccccc}
\toprule
\textbf{Group} & 3 & 4 & 5 & 7 & 9 & 10 & 11 & 12 & 13 & 14 & \textbf{Ave} \\
\midrule
Speech                    & 1.000 & 0.979 & 1.000 & 0.990 & 1.000 & 1.000 & 0.832 & 1.000 & 0.908 & 1.000 & 0.971 \\
Structure                 &  0.826 & 0.943 & 0.901 & 0.939 & 1.000 & 0.587 & 0.924 & 0.890 & 0.916 & 1.000 & 0.893 \\
Gesture (union)           & 0.578 & 0.540 & 0.638 & 0.507 & 0.568 & 0.542 & 0.526 & 0.378 & 0.452 & 0.222 & 0.495 \\
Gesture (intersection)    & 0.944 & 0.882 & 0.839 & 0.854 & 0.853 & 0.795 & 0.793 & 0.806 & 0.886 & 0.800 & 0.845 \\
\bottomrule
\end{tabular}}
\caption{Inter-annotator agreement (IAA) scores by modality across 10 groups.}
\label{fig:iaa-groups-modality}
\end{table*}


From the two annotation sets produced for each modality, inter-annotator agreement (IAA) was computed to assess reliability. For the speech and structure modalities, we used Cohen’s Kappa coefficient \cite{cohen1960coefficient} as the evaluation metric. In the speech modality, agreement was assessed on both the number and the propositional content of relational expressions extracted from speech acts. For the structure modality, agreement was calculated by comparing the human-annotated 3D structures, represented in $xyz$-coordinates along with the corresponding layer indices derived from the task video. Since action annotations were deterministically extracted from structure annotations, IAA over the structure annotations serves as validation of the action annotations as well. For the gesture modality, we report two scores. Annotators needed to record both when they saw a gesture occurring and what they interpreted it to mean. Since the two annotators may be more or less conservative with their identification of gesture {\it occurrence}, independent of interpretation, the two annotators may report different numbers of gestures in a video. This is similar to the problem of {\it mention} over- or under-annotation reported in coreference corpora \cite{bugert2021generalizing}. Therefore we report IAA scores over the union of the sets of gestures annotated by both annotators, and over the intersection only. For each annotator, following \citet{lai-etal-2024-encoding}, we concatenate all or some of their GAMRs into a single AMR: to calculate union scores we include all GAMRs, while for intersection scores we only include those gestures annotated by both annotators. We then calculate SMATCH F1 scores~\cite{cai2013smatch} between the two annotators' resulting AMRs.

Table~\ref{fig:iaa-groups-modality} presents the inter-annotator agreement (IAA) scores across three modalities for ten groups. Speech annotations exhibit consistently high levels of agreement, with an average Cohen’s Kappa score of 0.971. Structure annotations generally demonstrate high reliability, with many groups achieving strong consistency, though some variability is observed. This variability arises when the same block is annotated with different $xy$-coordinates by different annotators; for example, in Group 10 such discrepancies led to divergent representations of the same structure and consequently a lower agreement score. For the gesture modality, we see that IAA over the union is significantly lower than IAA over the intersection, due to the occurrence identification differences, although the numbers still signal moderate to substantial agreement (avg. of 0.495). IAA over the intersection is uniformly high (avg. 0.845), signaling reliable gesture annotation. Therefore we use the union of the reported annotations since we are concerned with gesture interpretation rather than mere occurrence identification, under the assumption that, when given the same set of identified gestures, the annotators display high agreement. We likewise assume that over-identified gesture occurrences that do not contribute to multimodal interpretation of meaning do not align with meaningful features in other modalities, and would therefore be filtered out in the next step (Sec.~\ref{ssec:xmodal}).

\subsection{Cross-Modality Alignment}
\label{ssec:xmodal}

In isolation, the individual modalities may not fully capture the instructions communicated by the directors, their expressed beliefs or personal grounding, or the builder's interpretation thereof. Therefore, we standardized and aligned the propositional content expressed through each primary modality to achieve a complete representation.


Each builder action (e.g., Builder: $(rs1,~on(gs2),~layer~2, D1’s~side)$) is parsed into a structured event specifying the acting block (top), the supporting block (base), the spatial relation (on or below), the layer number, and the speaker’s perspective. These structured action events establish mappings between actions and linguistic or gestural communications. For every speech or gesture interval, the pipeline retrieves temporally adjacent actions within configurable windows, enabling utterances such as “put the green on the red” and corresponding pointing gestures in the correct physical context.

Verbal and gestural propositions are normalized and symbolically grounded. Color-shape terms in speech (e.g., $RedShort$, $YellowLong$) are automatically replaced with concrete block identifiers (e.g., $rs2$, $yl1$) inferred from future action pairs. In gesture annotations, gesturer and gesture content are replaced with the participant ID and the nearest object, including its construction layer. 

All layer-aware propositions are then serialized into temporally aligned outputs for each modality and merged with emblematic gesture annotations that convey disagreement or confirmation. 

\section{Evaluation}
\label{sec:eval}

Our experimental setup evaluates the abilities of both (1) LLMs and (2) an axiomatic processing pipeline, at inferring what the collaboratively-built structure should look like, given the observable interaction between the participants, and at inferring their apparent shared beliefs. The state of the structure, as derived from the structure and action annotations, represents an objective observable ground truth that can be taken to be ultimately reflective of what the accepted common ground of the group is, after they have negotiated and disambiguated the placement of each block.

\subsection{Common Ground (CG) Calculation}
\label{ssec:axioms}


The temporally aligned multimodal propositions form the basis of our belief-tracking framework. For each speaker, {\it belief states} were annotated by two annotators over each dialogue, and they were encoded as to whether each relation is currently accepted, doubted, or negated. The dual annotations yielded an average inter-annotator agreement score of $\kappa = 0.809$. The annotated propositions, together with acceptance and doubt labels, are temporally aligned (Sec.~\ref{ssec:xmodal}). 

We then implemented a common-ground inference module that axiomatically predicts how shared knowledge and mutual belief evolve within and across turns. We specify the following axioms, inspired by \citet{bolander2014seeing,pustejovsky2021embodied} and \citet{grice1989studies} and grounded in a simplified model of evidence-based dynamic epistemic logic (EB-DEL) following \citet{van2014evidence} and \citet{pacuit2017neighborhood}, and used by \citet{khebour2024commongroundtrackingmultimodal} and \citet{pustejovsky2024lexical}:
\begin{enumerate}
\item \textbf{Seeing is Believing}: perceptual context directly updates belief;
\item \textbf{Acting is Believing}: embodied action reveals intention;
\item \textbf{Saying is Believing}: language communicates epistemic state.
\end{enumerate}
These axioms leverage the primary available modalities of language, gesture, and action, as available in the DPIP task. As belief states evolve based on these axioms, we incrementally construct and update sets of participants who share the same positive stance toward a given proposition. Whenever at least two individuals concurrently accept a relation, a CG(Common Ground) label (e.g., $CG_{\{D1,D2,Builder\}}: on(gs3,~rs2,~layer~2)$) is generated. Doubts temporarily remove the doubting participant from that CG(Common Ground) set, while negations delete the shared belief entirely—analogous to removing a physical block from the construction. Multi-clause acceptances and subset relations are also handled, allowing higher-level CG(Common Ground) formations to emerge from overlapping atomic agreements. This process yields a structured, dynamically updated representation of the group’s common ground under the above axiomatic assumptions.

Those CG calculations are then used to define the turn boundaries, where each turn represents a complete episode of belief formation. A turn is bounded by two successive updates in the shared common ground—the set of task-relevant beliefs mutually recognized by the group. Rows containing explicit common-ground (CG) annotations mark these boundaries, so that each turn spans the interval between consecutive CG labels, encompassing all intermediate multimodal events (verbal, gestural, and physical) that collectively constitute the evidential context leading to a belief update. The same turn boundaries were used to segment the dialogue sequences provided to the LLMs for evaluation, enabling direct comparison between human-annotated and model-predicted belief dynamics.

\subsection{Experiments}
\label{ssec:exp}

The following experiments were performed: 
\begin{enumerate}
    \item {\it Structure prediction from actions} ({\bf Action $\rightarrow$ Structure}): We input the annotated actions taken during each turn, and prompt an LLM to output the status of the structure after that turn, as the group proceeds toward task completion. This provides a strict measure of how actions can indicate {\it task progress}, and serves as a baseline to demonstrate how well LLMs can track actions taken on the board over time.
    \item {\it Structure prediction from aligned annotations} ({\bf Aligned $\rightarrow$ Structure}): \textcolor{black}{We input annotations from \textit{all} modalities without fusion for each turn, and prompt an LLM to output the status of the structure after that turn.} This measures how much the additional modalities that inform the block placements help or hurt LLMs' structure prediction ability.
    \item {\it Structure prediction from axiomatic CG} ({\bf CGC $\rightarrow$ Structure}): We run the CG inference (Sec.~\ref{ssec:axioms}) over the \textcolor{black}{aligned and merged multimodal} annotations to generate a predicted set of beliefs about block positions. These beliefs are then used to specify what the structure would be in the case that all the predicted beliefs were true. This demonstrates how well an axiomatic prediction of common ground matches the reality of what the group actually builds.
\end{enumerate}
Because each turn represents an episode of belief formation, sharing, and negotiation, and because the structure annotations demonstrate a high reliability, the physical board state after each turn represents an objective ``observable ground truth'' of the common ground between the group members, and so the outputs of these experiments are compared to the actual structure state.
\begin{enumerate}
  \setcounter{enumi}{3}
  \item {\it Common ground prediction from aligned annotations} ({\bf Aligned $\rightarrow$ CGC}): We input aligned annotations for each turn to an LLM and, rather than prompt it to predict the structure state, we prompt it to predict the {\it common ground} (shared belief set) of the group, and compare this output to the CG calculated axiomatically over the same annotations. This provides an indication of how much LLMs and an axiomatic approach overlap or differ in their predictions, and serves as an intrinsic evaluation of the CG calculation (CGC) axioms. 
\end{enumerate}



We conduct experiments using Qwen3-4B-Instruct-2507 \cite{qwen3} and Llama-3.2-3B-Instruct~\cite{dubey2024llama}, to assess multiple models on a limited compute budget. For a larger/stronger model comparison, we also evaluated GPT-5-mini~\cite{gpt-5}, and, when necessary, GPT-5~\cite{gpt-5}. The latter was employed when GPT-5-mini omitted outputs when performing structured reasoning or JSON generation.

Our primary metric is the Dice Similarity Coefficient (DSC; \citet{Srensen1948AMO,dice1945measures}), which measures the overlap between predicted and ground-truth sets with sensitivity to the size of the set. The Action $\rightarrow$ Structure task evaluates an LLM on its ability to produce each director's view as a 3x3 list of blocks, which is then translated to a set of configurational relations. For the Aligned $\rightarrow$ Structure prediction experiments, the LLM produces a set of spatial relations between blocks, whereas for the axiomatic CG prediction experiment, the LLM output is the set of predicted shared beliefs of the group. We report the \textit{Average} of DSC after each turn (reflecting local agreement on per-turn state changes) and DSC at the \textit{Global} dialogue level (agreement between the final predicted state and final actual state).

As mentioned in Sec.~\ref{ssec:data}, in the outlier group, Group 7, the participants did not successfully build a structure consistent with all three directors' side views. Thus, it provided an opportunity to study LLM performance and the contribution of multimodal annotations in a setting decoupled from task success as a completion state. In this case, since intermediate actions did \textit{not} lead to successful completion, we evaluate only on the aligned and axiomatic inputs (i.e., Experiments 2--4 as described above), to examine multimodal alignment and common ground on an example of task failure.

\section{Results and Analysis}
\label{sec:results}

\begin{table*}[h!]
\centering

\resizebox{\textwidth}{!}{
\begin{tabular}{lcccccccccccc}
\toprule
\textbf{Actions $\rightarrow$ Structure} & \textbf{Group} & 3 & 4 & 5 & 9 & 10 & 11 & 12 & 13 & 14 & \textbf{$\mu$} & \textbf{SD} \\
\midrule

\multirow{2}{*}{\quad \textbf{Llama 3.2-3B}} 
& Average               & 0.173 & 0.202 & 0.170 & 0.275 & 0.157 & 0.402 & 0.259 & 0.134 & 0.200 & 0.219 & 0.078 \\
& Global                & 0.000 & 0.037 & 0.056 & 0.056 & 0.000 & 0.148 & 0.074 & 0.037 & 0.037 & 0.049 & 0.042 \\
\cmidrule(lr){2-13}

\multirow{2}{*}{\quad \textbf{Qwen3-4B}} 
& Average               & 0.185 & 0.193 & 0.232 & 0.457 & 0.235  & 0.392 & 0.276 & 0.251 & 0.264 & 0.276 & 0.085 \\
& Global                & 0.037 & 0.000 & 0.148 & 0.185 & 0.074 & 0.148 & 0.148 & 0.037 & 0.185 & 0.107 & 0.066 \\
\cmidrule(lr){2-13}

\multirow{2}{*}{\quad \textbf{GPT-5-mini / GPT-5}} 
& Average               & 0.343 & 0.270 & 0.308 & 0.633 & 0.254 & 0.561 & 0.333 & 0.362 & 0.378 & 0.382 & 0.122 \\
& Global                & 0.133 & 0.148 & 0.000 & 0.482 & 0.000 & 0.222 & 0.000 & 0.000 & 0.444 & 0.159 & 0.180 \\
\midrule
\textbf{Aligned $\rightarrow$ Structure} & \textbf{Group} & 3 & 4 & 5 & 9 & 10 & 11 & 12 & 13 & 14 & \textbf{$\mu$} & \textbf{SD} \\
\midrule

\multirow{2}{*}{\quad \textbf{Llama 3.2-3B}} 
& Average & 0.110 & 0.097 & 0.155 & 0.198 & 0.105 & 0.099 & 0.080 & 0.106 & 0.115 & 0.118 & 0.036 \\
& Global  & 0.439 & 0.453 & 0.419 & 0.718 & 0.556 & 0.348 & 0.375 & 0.457 & 0.474 & 0.471 & 0.110 \\
\cmidrule(lr){2-13}

\multirow{2}{*}{\quad \textbf{Qwen3-4B}} 
& Average & 0.148 & 0.153 & 0.157 & 0.146 & 0.131 & 0.191 & 0.098 & 0.098 & 0.132 & 0.139 & 0.029 \\
& Global  & 0.811 & 0.704 & 0.545 & 0.903 & 0.667 & 0.703 & 0.541 & 0.516 & 0.621 & 0.668 & 0.130 \\
\cmidrule(lr){2-13}

\multirow{2}{*}{\quad \textbf{GPT-5-mini / GPT-5}} 
& Average & 0.030 & 0.000 & 0.000 & 0.075 & 0.017 & 0.025 & 0.010 & 0.065 & 0.046 & 0.029 & 0.027 \\
& Global  & 0.261 & 0.000 & 0.000 & 0.696 & 0.091 & 0.174 & 0.125 & 0.556 & 0.348 & 0.250 & 0.233 \\
\midrule
\textbf{CGC $\rightarrow$ Structure} & \textbf{Group} & 3 & 4 & 5 & 9 & 10 & 11 & 12 & 13 & 14 & \textbf{$\mu$} & \textbf{SD} \\
\midrule
& Average & 0.121 & 0.073 & 0.095 & 0.057 & 0.036 & 0.012 & 0.045 & 0.081 & 0.037 & 0.062 & 0.034 \\
& Global  & 0.519 & 0.364 & 0.455 & 0.889 & 0.357 & 0.239 & 0.197 & 0.167 & 0.271 & 0.369 & 0.215 \\
\midrule
\textbf{Aligned $\rightarrow$ CGC} & \textbf{Group} & 3 & 4 & 5 & 9 & 10 & 11 & 12 & 13 & 14 & \textbf{$\mu$} & \textbf{SD} \\
\midrule

\multirow{2}{*}{\quad \textbf{Llama 3.2-3B}} 
& Average & 0.167 & 0.242 & 0.400 & 0.185 & 0.000 & 0.016 & 0.000 & 0.098 & 0.153 & 0.140 & 0.131 \\
& Global  & 0.190 & 0.353 & 0.800 & 0.400 & 0.000 & 0.062 & 0.000 & 0.073 & 0.108 & 0.221 & 0.261 \\
\cmidrule(lr){2-13}

\multirow{2}{*}{\quad \textbf{Qwen3-4B}} 
& Average & 0.167 & 0.000 & 0.000 & 0.074 & 0.083 & 0.000 & 0.104 & 0.275 & 0.334 & 0.115 & 0.121 \\
& Global  & 0.286 & 0.000 & 0.000 & 0.182 & 0.250 & 0.000 & 0.073 & 0.296 & 0.158 & 0.138 & 0.124 \\
\cmidrule(lr){2-13}

\multirow{2}{*}{\quad \textbf{GPT-5-mini / GPT-5}} 
& Average & 0.375 & 0.515 & 0.200 & 0.333 & 0.278 & 0.000 & 0.292 & 0.407 & 0.195 & 0.288 & 0.148 \\
& Global & 0.435 & 0.571 & 0.500 & 0.429 & 0.727 & 0.000 & 0.214 & 0.281 & 0.108 & 0.363 & 0.232 \\
\bottomrule
\end{tabular}}
\vspace*{-2mm}
\caption{Results in four experimental conditions for all models, methods, and metrics (DSC).}
\label{tab:all-models-results-structure-prediction}
\vspace*{-2mm}
\end{table*}

\begin{table}[h!]
\centering
\resizebox{\linewidth}{!}{
\begin{tabular}{lccc}
\toprule
\multirow{2}{*}{\textbf{Aligned $\rightarrow$ Structure}} 
& \multirow{2}{*}{\begin{tabular}[c]{@{}c@{}}\textbf{Llama}\\\textbf{3.2-3B}\end{tabular}} 
& \multirow{2}{*}{\begin{tabular}[c]{@{}c@{}}\textbf{Qwen3}\\\textbf{-4B}\end{tabular}} 
& \multirow{2}{*}{\begin{tabular}[c]{@{}c@{}}\textbf{GPT-5-mini}\\\textbf{/ GPT-5}\end{tabular}} \\
& & & \\
\midrule
\quad Average & 0.040 & 0.066 & 0.007 \\
\quad Global  & 0.182 & 0.298 & 0.055 \\
\midrule
\multicolumn{4}{l}{\textbf{CGC $\rightarrow$ Structure}} \\
\midrule
\quad Average & \multicolumn{3}{c}{0.250} \\
\quad Global  & \multicolumn{3}{c}{0.226} \\
\midrule
\multirow{2}{*}{\textbf{Aligned $\rightarrow$ CGC}} 
& \multirow{2}{*}{\begin{tabular}[c]{@{}c@{}}\textbf{Llama}\\\textbf{3.2-3B}\end{tabular}} 
& \multirow{2}{*}{\begin{tabular}[c]{@{}c@{}}\textbf{Qwen3}\\\textbf{-4B}\end{tabular}} 
& \multirow{2}{*}{\begin{tabular}[c]{@{}c@{}}\textbf{GPT-5-mini}\\\textbf{/ GPT-5}\end{tabular}} \\
& & & \\
\midrule
\quad Average & 0.334 & 1.000 & 1.000 \\
\quad Global  & 0.500 & 1.000 & 1.000 \\
\bottomrule
\end{tabular}
}
\vspace*{-2mm}
\caption{DSC Results in three experimental conditions for outlier group, Group 7.}
\label{tab:g7-prediction}
\vspace*{-4mm}
\end{table}

Table~\ref{tab:all-models-results-structure-prediction} shows results in all four experiments for all methods and metrics over 9 of the 10 annotated DPIP groups.


In the first experiment, {\it Actions $\rightarrow$ Structure}, GPT-5 is the best performer, with a statistically significant advantage over the next-best model, Qwen, according to group-wise Average DSCs ($p\approx0.004$ by a Wilcoxon signed-rank test). However, Global DSC scores are worse for all models than Average DSCs, suggesting that when given only information about actions, LLMs are better at inferring structure at the turn level than over entire dialogues. This aligns with previous results showing that LLMs and common methods with which they are aligned for generation struggle with longer multi-turn interactions and reasoning \cite{zhou2024archer}.

In the {\it Actions $\rightarrow$ Structure} experiment, LLMs were provided with only one modality, and so it can serve as an approximate baseline to assess how adding other modalities helps or hurts inference ability. When provided with aligned annotations across all available modalities ({\it Aligned $\rightarrow$ Structure}),  Qwen is the top performer (statistically significant advantage over nearest competitor Llama, $p\approx0.004$). 
Here, in all cases between the two experiments, Average step-wise DSC goes down but Global DSC goes up.  This may reflect added context via other modalities.
If so, the additional context seems more useful when looking at entire dialogues together rather than segmented turns, suggesting that with sufficient context, additional modalities help, but without it, they simply add noise.
Interestingly in this case, GPT-5-mini/GPT-5 goes from the best-performing LLM to the worst, predicting exactly none of the correct structure for Groups 4 and 5. It seems surprisingly bad at processing the DPIP task data despite its formidable reputation, and also has a very high standard deviation compared to raw per-group scores. 


{\it CGC $\rightarrow$ Structure}, the deterministic axiomatic approach to inferring common ground, is sometimes surprisingly effective at predicting the structure state. CG axioms achieve higher mean Average and Global DSC scores at structure prediction than GPT-5-mini does when given access to all modalities (non-significant medium difference, $p\approx0.2$).

Group 9 is consistently among the highest-scoring groups, and seems to be relatively easy for both LLMs and CGC to process.

{\it Aligned $\rightarrow$ CGC} computes the overlap between LLMs' predictions of common ground and common ground calculated axiomatically. We find that overall overlap is quite low, indicating the LLMs frequently infer different belief states from multimodal dialogues than are calculated axiomatically. In many cases the DSC is 0, indicating completely disjunct sets predicted by the two methods. Interestingly, on Group 9, on which both LLMs and CGC achieve high global DSC on structure prediction, the LLM/CGC overlap never exceeds 0.5, indicating that even on an apparently ``easy" dialogue in this task, LLMs and CG axioms are retrieving at least very different aspects of the interaction.

Overall there is a high variance in performance across groups, similar to what is observed by \citet{khebour2024commongroundtrackingmultimodal} in their common ground tracking task in a multimodal dialogue. This convergent result further highlights the diverse ways in which collaborative groups express belief convergence and epistemic positioning.

Table~\ref{tab:g7-prediction} shows performance on Group 7, the outlier group that failed to complete the task correctly. In the {\it Aligned $\rightarrow$ Structure} and {\it CGC $\rightarrow$ Structure} experiments, we see results that fall within the distribution of results for the other groups. However, a very interesting result emerged when evaluating {\it Aligned $\rightarrow$ CGC}: both Qwen and GPT perfectly infer the axiomatically-calculated common ground. This result seems sharply divergent from expectations of what should be a harder inference problem (i.e., common ground in a group that fails at the task). \textcolor{black}{However, a closer examination of the group dynamics explains this outcome: widespread confusion about the task, goals, and available information leaves little shared common ground that can be axiomatically extracted.}
The LLMs, when presented with the aligned annotations, infer small or even empty belief sets, which matches the lack of common ground displayed by the group. They appear able to correctly detect when there is a {\it lack} of common ground displayed in the group dialogues and annotations. Conversely, as indicated by {\it Aligned $\rightarrow$ CGC} over other groups, when there exists substantive common ground in a group, as calculated over the annotations, SOTA LLMs remained challenged by the task of inferring what its contents are.

\section{Conclusion}
\label{sec:conc}

In this paper, we introduced a challenging new task, the Distributed Partial Information Puzzle (DPIP), realized as a collaborative construction task under partially-observable conditions. This task and data uniquely combines a multiparty, co-situated interaction with rich belief dynamics induced by the partial information setting, which simulates bringing together teams composed of members with different backgrounds and expertise. We performed rich multimodal annotations over the data, and evaluated 3 modern LLMs as well as an axiomatic belief extraction pipeline on the task of predicting the structure being built based on the participants' utterances, gestures, and/or actions in context. Our results show how challenging this task setting and data is for the modeling of belief state, group dynamics, and task progressing by state of the art systems, establishing a challenging new task and benchmark for multimodal dialogue research, including common ground tracking \cite{khebour2024commongroundtrackingmultimodal,tu2024dense,vanderhoeven2025trace}, but also for tasks such as modeling theory of mind~\cite{bara-etal-2021-mindcraft,sileo2023mindgames} in small groups, or spatiotemporal reasoning. 

\section*{Limitations}

Although 3 camera angles were recorded during collection, annotations were conducted from a single camera angle for each session (the central of the three). As a result, it is possible that at times part of the structure could be occluded from view, e.g., when participants placed blocks behind existing ones. This could potentially introduce uncertainty while annotating the spatial relationships of the blocks. 

\textcolor{black}{The reliability of identifying gesture occurrences remains a significant methodological challenge within the field. While we currently evaluate inter-annotator agreement (IAA) based on the union of annotated gesture sets, this approach may be refined through the application of tools such as Staccato \cite{lucking2011assessing}.}

The fully annotated data which we evaluated on constitutes only a subset of the total data collected. We annotated and reported statistics and results on 10 groups out of a full 33. The total length of all 33 videos is 19 hours, 46 minutes, and 27 seconds in length. As mentioned in Sec.~\ref{ssec:dpip}, sessions contain a variant of the task that is less structured (larger footprint, gaps allowed in the structure, greater variety of block types). Our results on the 10 groups in the simpler of the two tasks already showcase the challenge of modeling and tracking common ground in the DPIP Lego task, and the less restrictive task variant is likely to be more challenging still. 

\section*{Ethical Considerations}

Collaborative construction tasks like Lego building exemplify creative construction as a developmental and epistemic practice extending beyond childhood. From a developmental ethics perspective, such tasks can foster autonomous imagination and collaborative sense-making, as participants negotiate meaning, share limited resources, and construct shared models of understanding through tangible forms. In our study, participants were provided with visual stimuli, pictures of target structures but no verbal or procedural instructions, compelling them to infer, coordinate, and reconstruct spatial relations through embodied reasoning and dialogue. Whereas this openness exposes the limits and frictions of creative collaboration, interpretive freedom may lead to conflict, misalignment, or the dominance of certain voices, while material constraints and the standardized logic of Lego pieces channel imagination into predefined geometries. What appears as freedom of construction thus may carry subtle forms of restriction and hierarchy, reminding us that even creative play may reproduce systems of order, control, and negotiation that mirror the social and epistemic structures from which it emerges.

Our data collection took place under an instantiated human subjects research protocol that was reviewed and approved by Colorado State University and Brandeis University Institutional Review Boards (IRB), as well as human subjects research protection offices at the agencies that funded the data collection. Participants consented to be recorded and were fairly compensated for their time (Sec.~\ref{ssec:data}). Training and evaluating AI models over recorded data containing human likenesses risks exposing participants to invasion of privacy, however in our experiments, no video, raw audio, or human likeness data was sent to an AI model (e.g., GPT-5), only transcripts and annotations.

\section*{Acknowledgements}
We are grateful to Tarun Varma Buddaraju, Jack Fitzgerald, Sai Kiran Ganesh Kumar, Sai Shruthi Garlapati, Carine Graff, Shamitha Gowra, Huma Jamil, Changsoo Jung, Ibrahim Khebour, Maniteja Vallala, Yangyang Chen,  and Marc Verhagen for their valuable assistance in the data collection and annotation process, and to Bruce Draper and Nathaniel Blanchard for additional work on the task development. We would also like to thank the anonymous reviewers whose valuable feedback helped improved the quality of the final copy of this manuscript.
This material is based in part upon work supported by Other Transaction award HR00112490377 from the U.S. Defense Advanced Research Projects Agency (DARPA) Friction for Accountability in Conversational Transactions (FACT) program, the U.S. National Science Foundation (NSF) under award DRL 2454151 (Institute for Student-AI Teaming), and by award W911NF-25-1-0096 from the U.S. Army Research Office (ARO). The views and conclusions contained in this document are those of the authors and should not be interpreted as representing the official policies, either expressed or implied, of the U.S. Government.

\nocite{*}
\section*{Bibliographical References}
\label{sec:reference}

\bibliographystyle{lrec2026-natbib}
\bibliography{lrec2026-example}


\end{document}